\pdfoutput=1

\documentclass[11pt]{article}

\usepackage[preprint]{acl}

\usepackage{times}
\usepackage{latexsym}

\usepackage[T1]{fontenc}

\usepackage[utf8]{inputenc}

\usepackage{microtype}

\usepackage{inconsolata}

\usepackage{graphicx}

\usepackage{amsmath}
\usepackage{multicol}
\usepackage{multirow}
\usepackage{algorithm}
\usepackage{algorithmic}
\usepackage{subfig}
\usepackage{graphicx}
\usepackage{booktabs}
\usepackage{marvosym}
\usepackage{url}

\title{Chip-Tuning: Classify Before Language Models Say}

\author{Fangwei Zhu$^{1}$\thanks{Work done during internship at Tencent QQ, as a part of QQ MLLM project.\\\indent ~ \textsuperscript{\Letter}Corresponding author.\\\indent ~ $^\Diamond$Project leader of QQ MLLM project.}, Dian Li$^{2\Diamond}$\textsuperscript{\Letter}, Jiajun Huang$^2$, Gang Liu$^2$, Hui Wang$^2$, Zhifang Sui$^{1}$\textsuperscript{\Letter} \\
  $^1$National Key Laboratory for Multimedia Information Processing, Peking University \\
  $^2$Tencent QQ \\
  \texttt{zhufangwei2022@stu.pku.edu.cn} \\
  \texttt{\{goodli, kakwanhuang, sinbadliu, joltwang\}@tencent.com} \\
  \texttt{szf@pku.edu.cn} \\
}

\begin{document}
\maketitle
\begin{abstract}
The rapid development in the performance of large language models (LLMs) is accompanied by the escalation of model size, leading to the increasing cost of model training and inference.
Previous research has discovered that certain layers in LLMs exhibit redundancy, and removing these layers brings only marginal loss in model performance.
In this paper, we adopt the probing technique to explain the layer redundancy in LLMs and demonstrate that language models can be effectively pruned with probing classifiers.
We propose chip-tuning, a simple and effective structured pruning framework specialized for classification problems.
Chip-tuning attaches tiny probing classifiers named chips to different layers of LLMs, and trains chips with the backbone model frozen.
After selecting a chip for classification, all layers subsequent to the attached layer could be removed with marginal performance loss.
Experimental results on various LLMs and datasets demonstrate that chip-tuning significantly outperforms previous state-of-the-art baselines in both accuracy and pruning ratio, achieving a pruning ratio of up to 50\%.
We also find that chip-tuning could be applied on multimodal models, and could be combined with model finetuning, proving its excellent compatibility.
Our code is available at \url{https://github.com/QQ-MM/ChipTuning}.
\end{abstract}

\section{Introduction}

Large language models (LLMs) have experienced rapid development in recent years, achieving surprising success in various domains.
Researchers have been scaling up the size of language models to pursue better performance, just as the scaling law~\cite{kaplan2020scaling} suggests.
However, the increasing size of models leads to massive computational costs, posing a challenge to practical deployment and usage.

Model compression techniques have since been proposed as a solution to relieving computational stress, which would assist in the deployment of large models.
Different approaches have been explored to compress language models into more compact versions, including quantization~\cite{liu2021post, dettmers2022gpt3, dettmers2024qlora}, knowledge distillation~\cite{gou2021knowledge, gu2023knowledge, ko2024distillm} and pruning~\cite{ma2023llm, yang2024laco, ashkboos2024slicegpt, men2024shortgpt}.

Relevant research~\cite{men2024shortgpt} reveals that a fair portion of parameters in large language models are redundant, and removing these parameters would not bring severe damage to the performance of models.
Based on the observation, different methods have been designed to identify and remove redundant parameters from LLMs, like layer merging~\cite{yang2024laco}, width compression~\cite{ashkboos2024slicegpt}, layer removal~\cite{men2024shortgpt} and component removal~\cite{ma2023llm}.
These methods maintain the majority of performance, proving the feasibility of model pruning.

Research on model interpretability has shown evidence that language models may develop internal representations for various features like color~\cite{patel2022mapping}, truthfulness~\cite{burns2022discovering}, chessboard states~\cite{nanda2023emergent}, numbers~\cite{zhu2024language} or even abstract concepts like code errors~\cite{templeton2024scaling}.
These features typically start to form on middle layers and will be carried to subsequent layers~\cite{stolfo2023mechanistic}.
More interestingly, many of these features can be read out by probing techniques~\cite{belinkov2022probing}, in the way of training simple classifiers.

Inspired by the discovery that removing late layers of LLMs does not heavily impair network functionality~\cite{men2024shortgpt}, we hypothesize that the critical features for solving certain problems may begin to form on intermediate layers of LLMs.
By probing these necessary features on intermediate layers, we can safely prune subsequent layers with marginal performance loss.

We observe that previous research mainly aimed to build a general pruned model that can be directly applied to various downstream tasks.
Based on the intuition that different tasks require different subsets of features, we hypothesize that pruning on specific tasks instead of pruning for a general model would yield better results, as the model could better focus on the related features.

In this paper, we introduce \textbf{chip-tuning}, a simple and effective structured pruning framework specialized for classification tasks.
For a given classification task, we attach probing classifiers named chips to each layer of the language model, and train these classifiers to probe the final classification results from intermediate hidden states.
After training, we can then select a chip with satisfactory accuracy, and prune all layers subsequent to the chip to obtain a more compact model for the task.
The parameters of the backbone model are frozen throughout the whole process and will not introduce any additional computation cost.

We apply chip-tuning to language models with different sizes and families and observe their performance on various classification tasks.
Compared with previous pruning methods, chip-tuning demonstrates better performance on classification tasks, and enables more radical pruning that reduces the parameters of models by up to 50\% with marginal loss in performance.
Additional experiments show that chip-tuning is also compatible with multimodal large language models (MLLMs) and other finetuning methods.

The main contributions of our paper can be summarized as:
\begin{itemize}
    \item We propose chip-tuning, a pruning framework for classification tasks that trains probing classifiers attached to certain layers of language models. By removing layers subsequent to the selected classifier, we can effectively reduce the size of the models.
    \item We conduct experiments on different benchmarks, experimental results show that Chip-tuning is able to maintain the performance while reducing the size of models by up to 50\%, much outperforming previous state-of-the-art baselines.
    \item We evaluate chip-tuning on multimodal models and finetuned models, whose results prove the excellent compatibility of chip-tuning.
\end{itemize}

\begin{figure*}
    \centering
    \includegraphics[width=0.9\linewidth]{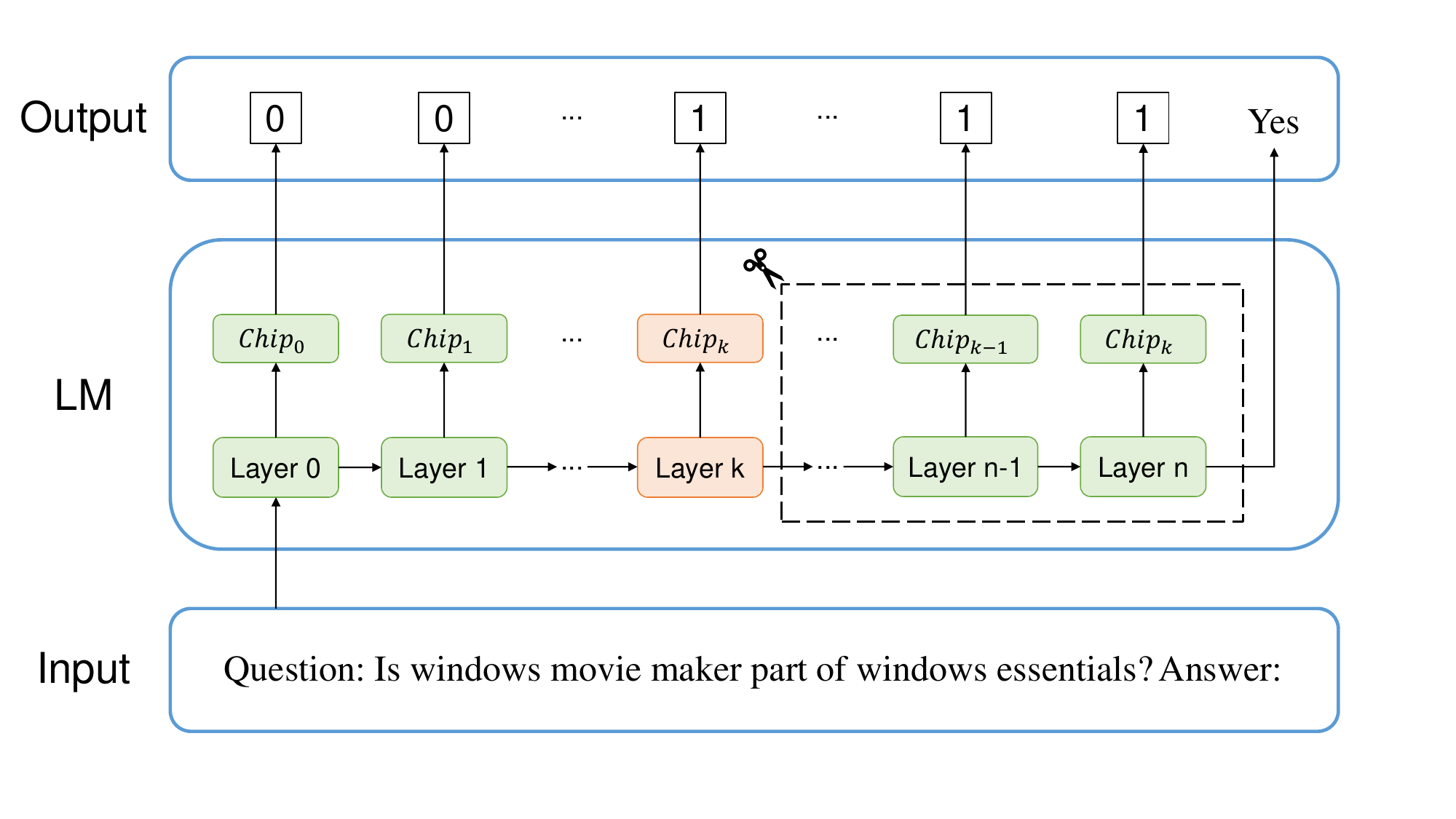}
    \caption{The overall structure of chip-tuning. After selecting a certain chip attached to the k-th layer, subsequent layers in the language model can be safely pruned with marginal influence on model performance. In training, only the parameters of chips are trainable and the backbone model is frozen.}
    \label{fig:datapath}
\end{figure*}

\section{Related Work}
\paragraph{Network Pruning.}
With the growth in the size of language models, the pruning technique has been proposed to eliminate unnecessary weights or structures in language models, thus accelerating language models.
The pruning methods can be generally categorized into two types: unstructured pruning and structured pruning.

Unstructured pruning methods focus on the level of individual weights, which try to speed up models by increasing the sparsity level of model weights.
SparseGPT~\cite{frantar2023sparsegpt} reduces the pruning problem to layer-wise sparse regression and incrementally prunes each column in the weight matrix with a sequence of Hessian inverses.
Wanda~\cite{sun2023simple} enhances the magnitude pruning approach with input activation norms, effectively reducing the complexity of pruning algorithms.
RIA~\cite{zhang2024plug} notices that previous methods tend to prune away entire channels of network weights, and mitigates the issue by jointly considering input and output channels.

Structured pruning methods operate at the level of network structures instead, which compress language models by removing redundant model components.
LLMPruner~\cite{ma2023llm} employs gradient information as a reference to remove non-critical structures.
SliceGPT~\cite{ashkboos2024slicegpt} removes rows or columns corresponding to small principal components in the weight matrix to achieve smaller weight matrices.
LaCo~\cite{yang2024laco} proposes the layer collapse algorithm, which merges adjacent layers while ensuring the representation similarity on few-shot calibration examples.
ShortGPT~\cite{men2024shortgpt} finds that deep layers of language models are not as effective as expected, and proposes the block importance metric to identify and remove redundant layers.
BlockPruner~\cite{zhong2024blockpruner} decomposes each Transformer layer into two minimal residual blocks and performs fine-grained block pruning to avoid significant performance loss.

\paragraph{Probing Language Models.}
The impressive capability of language models raises the hypothesis that language models have gone beyond mere memorization of surface correlations.
Instead, they may learn the principles behind the training data and develop internal representations for features~\cite{belinkov2022probing}. 
A wide variety of features have been detected in the hidden state of language models like color~\cite{patel2022mapping} and truthfulness~\cite{burns2022discovering}.

Probing is a widely adopted technique to associate internal representations with external properties~\cite{belinkov2022probing}.
By training a simple classifier on model representations that predicts a given property, we can read out various features before language models generate their final outputs.
With a simple linear classifier, probing is able to extract complex features like board game states~\cite{nanda2023emergent}, entity properties~\cite{li2021implicit}, and spatial information~\cite{gurnee2023language}.

Recently, ~\citet{tao2024probing} finds that probing classifiers are also able to extract cross-modal information from multimodal large language models.
\citet{zhang2024visually} further reveals that probing could achieve better performance on image classification tasks even than directly finetuning the backbone models. 

An interesting discovery is that probing classifiers sometimes achieve the best performance at intermediate layers, rather than early or late layers~\cite{zhu2024language}.
A hypothesis is that late layers focus more on local features related to the next token prediction, while intermediate layers gather information in the input text and thus contain more global information~\cite{stolfo2023mechanistic}.

\section{Methodology}
We illustrate the structure of the chip-tuning framework in Figure \ref{fig:datapath}.
The framework first inserts simple probing classifiers named chips to different layers of language models, and then solely trains the chips on task-specific training data.

Finally, we can select the chip on a fixed layer or with other strategies (see Section \ref{ssec:chip_selection}), and layers subsequent to the attached layer will be removed.

\subsection{Chips}
A language model with the decoder-only structure consists of $L$ transformer layers.
At every token position $t$, each transformer layer $l$ takes previous partial sequence $x_{\leq t}$ as input and outputs new hidden states $x_{t}^{l}$.

As discovered by previous research on probing, the hidden states of intermediate layers may contain rich features that can be read out by probing classifiers.
Chips are simple probing classifiers that try to predict the classification label $y$ from certain hidden states $x_{t}^{l}$.

We use two types of chips in our experiments: linear chips $p_{L}$ and two-layer perceptron (2xMLP) chips $p_{M}$, whose function could be notated as:

\begin{equation*}
    p_{L}(x_{t}^{l}) = \text{softmax}(Wx_{t}^{l} + b)
\end{equation*}
\begin{equation*}
    p_{M}(x_{t}^{l}) = \text{softmax}(W_{1} \text{ReLU}(W_{2} x_{t}^{l} + b_{2}) + b_{1}
\end{equation*}
where $W$, $W_1$, $W_2$, $b$, $b_1$ and $b_2$ are trainable parameters.

For simplicity, we take the hidden state at the last token position (i.e. $t = -1$) as the input vector of chips.

\subsection{Training}
As the optimal layer $l^*$ for classification chips is initially unknown, we attach a chip $p^{l}$ to every layer $l$ of the language model, and train these chips simultaneously with standard cross-entropy loss:
\begin{align*}
    \mathcal{L}^{l} &= y\log p(x^{l}_{t}) + (1-y) \log (1-p(x^{l}_{t})) \\
    \mathcal{L} &= \sum_{i=0}^{L} L^{l}
\end{align*}

Note that the parameters in the backbone language model are frozen in the training process, and only the weights of chips would be updated.

\subsection{Layer Removal and Inference}
We use the straightforward layer removal method to reduce the size of language models.
After selecting chip $p^{l}$ at layer $l$ as the classification chip, we simply remove all layers after layer $l$ to obtain a smaller model.

Namely, with chip $p^{l}$ at layer $l$ finally selected, the pruned model would function as follows:
\begin{algorithm}[!h]
    \renewcommand{\algorithmicrequire}{\textbf{Input:}}
    \renewcommand{\algorithmicensure}{\textbf{Output:}}
    \caption{Inference with Chips}
    \begin{algorithmic}[1]
        \REQUIRE Language model $M$ with $N$ layers $L_0, L_1, \ldots, L_{N-1}$, selected chip $p^{l}$ at layer $l$, input embedding $x^{-1}$;
    
        \ENSURE Classification prediction $y$;
    
        \FORALL{$i = 0, 1, 2, \ldots, l$}
            \STATE $x^{i} = L_{i}(x^{i-1})$
        \ENDFOR
        \STATE $y = \arg \max(p^{l}(x^{i}_{t}))$
    \end{algorithmic}
\end{algorithm}

Chip-tuning also supports inference with multiple chips $p_0, p_1, \ldots, p_{n-1}$ for different tasks.
Assuming that the deepest layer with chips attached is $l_m = \max(l_0, l_1, \ldots, l_{n-1})$, layers subsequent to $l_m$ will be pruned.

\section{Experiments}
\subsection{Experimental Setup}

\paragraph{Benchmarks.}
We select 4 distinct benchmarks on natural language processing with the form of multi-choice for evaluation: 
MMLU~\cite{hendrycks2020measuring}, Race~\cite{lai2017race}, BoolQ~\cite{clark2019boolq} and C3~\cite{sun2020investigating}. 

Furthermore, we introduce three image classification datasets to test the effectiveness of chip-tuning on multimodal large language models (MLLMs): Flowers102~\cite{nilsback2008automated}, StanfordCars~\cite{krause20133d}, and Caltech101~\cite{fei2004learning}, each containing 102, 196, and 101 classes respectively.

\paragraph{Models.}
Following previous work~\cite{men2024shortgpt}, we choose 2 model series to evaluate the effectiveness of chip-tuning: Llama2~\cite{touvron2023llama}, Baichuan2~\cite{yang2023baichuan}, which share similar decoder-only transformer structure. 
We use the 7B and 13B versions of Llama2 and Baichuan2 for experiments.
For multimodal large language models, we use the 7B and 13B versions of LLaVA-1.5~\cite{liu2023improvedllava} as the backbone model.

Due to memory constraints, we run 13B models under the precision of 16-bit (fp16) instead of 32-bit (fp32).

\paragraph{Baselines.}
We compare our method with several structured pruning methods: 
\textbf{LLMPruner}~\cite{ma2023llm} removes non-critical coupled structures on the basis of gradient information.
\textbf{SliceGPT}~\cite{ashkboos2024slicegpt} replaces weight matrices with smaller matrices by retaining principal components.
\textbf{LaCo}~\cite{yang2024laco} merges the layer in language models from deep to shallow, and sets a threshold to prevent excessive merging.
\textbf{ShortGPT}~\cite{men2024shortgpt} removes redundant layers according to their proposed Block Influence metric, a variant of cosine similarity.

\paragraph{Settings.}
For each benchmark, we use at most $20,000$ training data in the corresponding training split of the benchmark to train our chips.
The chips are trained with a batch size of $1$ for $1$ epoch.
We use a learning rate of $1\times 10^{-5}$ for our experiments, and set the hidden dimension of MLP chips to 256.
All experiments are conducted on a single NVIDIA A100 40GB GPU.

For 7B models, we select chips at layer 20 as classification chips; for 13B models, we select chips at layer 25 as classification chips.
These settings are equal to the prune ratio of $34.4\%$ and $35.0\%$, respectively.

\begin{table*}[t]
    \centering
    \scalebox{0.9}{
        \begin{tabular}{ccccccccc}
		\toprule
        Model & Method & Ratio (\%) & BoolQ & Race-H & Race-M & C3 & MMLU & Avg. Score \\  
		\midrule
		\multirow{8}{*}{Llama2-7B} & Dense & 0.00\% & 71.62 & 35.71 & 34.19 & 43.56 & 45.39 & 46.09 \\
		& LLMPrun. & 27.0\% & 55.20 & 22.56 & 22.35 & 25.64 & 23.33 & 29.82 \\
		& SliceGPT  & 26.4\% & 38.32 & 21.07 & 21.66 & 39.78 & 28.92 & 29.95 \\
		& LaCo  & 27.1\% & 64.07 & 22.61 & 23.61 & 39.67 & 26.45 & 35.28 \\
		& ShortGPT  & 27.1\% & 74.71 & 32.25 & 35.17 & 39.62 & 43.96 & 45.14 \\
        & CT (Lin.) & 34.4\% & 79.05 & 47.91 & 53.69 & 48.93 & 44.89 & 54.89 \\
        & CT (MLP) & 34.4\% & 76.01 & 49.43 & 53.90 & 53.80 & 45.07 & 55.64 \\
        & CT (Max) & \textasciitilde 40\% & \textbf{79.48} & \textbf{50.34} & \textbf{54.74} & \textbf{54.35} & \textbf{45.49} & \textbf{56.88} \\
		\midrule
		\multirow{8}{*}{Llama2-13B} & Dense & 0.00\% & 82.39 & 57.95 & 60.38 & 47.51 & 55.00 & 60.65 \\
		& LLMPrun. & 24.4\% & 56.42 & 22.47 & 22.08 & 32.33 & 25.21 & 31.70 \\
		& SliceGPT & 23.6\% & 37.86 & 23.41 & 24.03 & 41.92 & 37.14 & 32.87 \\
		& LaCo & 24.6\% & 63.98 & 54.49 & 56.55 & 44.93 & 45.93 & 53.18 \\
		& ShortGPT & 24.6\% & 62.48	& 58.35 & 60.17 & 46.90 & \textbf{54.69} & 56.52 \\
        & CT (Lin.) & 35.0\% & 78.23 & 62.04 & 67.06 & 68.21 & 52.79 & 65.67 \\
        & CT (MLP) & 35.0\% & 75.81 & 62.52 & 67.13 & 68.00 & 52.95 & 65.28 \\
        & CT (Max) & \textasciitilde 50\% & \textbf{79.76} & \textbf{63.29} & \textbf{68.04} & \textbf{69.39} & 53.41 & \textbf{66.78} \\
		\midrule
		\multirow{8}{*}{Baichuan2-7B} & Dense & 0.00\% & 74.10 & 26.96 & 24.09 & 64.55 & 53.87 & 48.71 \\
		& LLMPrun. & 24.2\% & 61.19 & 21.96 & 22.28 & 41.64 & 24.93 & 34.40 \\
		& SliceGPT & 22.2\% & 39.30 & 23.53 & 22.49 & 26.58 & 25.18 & 27.42 \\
		& LaCo & 24.2\% & 56.15 & 28.99 & 27.72 & 50.85 & 31.53 & 39.05 \\
		& ShortGPT & 24.2\% & 67.83 & 53.26 & 46.76 & 56.33 & 45.77 & 53.99 \\
        & CT (Lin.) & 34.4\% & 72.78 & 62.69 & 66.85 & 75.47 & 51.09 & 65.78 \\
        & CT (MLP) & 34.4\% & 73.12 & 63.52 & 67.13 & 76.36 & 50.95 & 66.22 \\
        & CT (Max) & \textasciitilde 40\% & \textbf{74.68} & \textbf{64.04} & \textbf{68.38} & \textbf{76.36} & \textbf{51.22} & \textbf{66.94} \\
		\midrule
		\multirow{8}{*}{Baichuan2-13B} & Dense & 0.00\% & 77.89 & 67.27 & 68.94 & 65.64 & 59.50 & 67.85 \\
		& LLMPrun. & 24.3\% & 56.54 & 21.17 & 21.61 & 39.89 & 23.19 & 32.48 \\
		& SliceGPT & 22.8\% & 37.83 & 21.56 & 21.52 & 24.99 & 22.95 & 25.77 \\
		& LaCo & 24.7\% & 62.35 & 56.92 & 57.80 & 61.10 & 51.35 & 57.90 \\
		& ShortGPT & 24.7\% & 62.54 & 55.77 & 56.41 & 60.16 & 52.11 & 57.40 \\
        & CT (Lin.) & 35.0\% & 77.77 & 73.04 & 77.44 & 80.84 & 56.88 & 73.19 \\
        & CT (MLP) & 35.0\% & 76.88 & 73.87 & 77.44 & 81.81 & 56.66 & 73.33 \\
        & CT (Max) & \textasciitilde 50\% & \textbf{78.84} & \textbf{75.04} & \textbf{79.11} & \textbf{81.89} & \textbf{56.96} & \textbf{74.37} \\
		\bottomrule
	\end{tabular}
 }
	\caption{Comparison of pruning methods on natural language benchmarks. CT refers to chip-tuning (our method). The results of LLMPrun., SliceGPT, LaCo, and ShortGPT are reported from ShortGPT~\cite{men2024shortgpt}. }
	\label{tab:llm_comparison_all}
\end{table*}

\begin{table*}[t]
	\setlength{\tabcolsep}{2.2pt}
	\centering
	\begin{tabular}{ccccccc}
		\toprule
		Model & Method & Ratio(\%) & Flowers102 & StanfordCars & Caltech101 & Avg. Score \\  
		\midrule
		\multirow{5}{*}{Llava1.5-7B} & Raw & 0.00\% & 5.9 & 0.0 & 47.1 & 17.67 \\
        & w/ Label & 0.00\% & 10.2 & 0.0 & 62.1 & 24.10 \\
        & CT (Lin.) & 34.4\% & 91.28 & 60.98 & 92.24 & 81.50 \\
        & CT (MLP) & 34.4\% & 88.70 & 0.85 & 91.52 & 60.36 \\
        & CT (Max) & \textasciitilde 20\% & \textbf{94.00} & \textbf{70.95} & \textbf{92.24} & \textbf{85.73} \\
		\midrule
		\multirow{5}{*}{Llava1.5-13B} & Raw & 0.00\% & 5.3 & 0.0 & 49.9 & 18.4 \\
        & w/ Label & 0.00\% & 7.2 & 0.1 & 70.9 & 26.07 \\
        & CT (Lin.) & 50.0\% & 91.46 & 48.63 & 91.70 & 77.26 \\
        & CT (MLP) & 50.0\% & 85.93 & 0.85 & 90.42 & 59.07 \\
        & CT (Max) & \textasciitilde 50\% & \textbf{93.06} & \textbf{71.52} & \textbf{92.39} & \textbf{85.66} \\
		\bottomrule
	\end{tabular}
	\caption{Comparison of pruning methods on image classification benchmarks. CT refers to chip-tuning (our method). The results of dense models are reported from \citet{zhang2024visually}.}
	\label{tab:mllm_comparison_all}
\end{table*}

\subsection{Main Experiment Results}
To evaluate the effectiveness of chip-tuning, we conduct experiments on multi-choice style benchmarks commonly used for large language model evaluation.
The experimental results are demonstrated in Table \ref{tab:llm_comparison_all}.
\footnote{We report the result of finetuning pruned baseline models with the same data used by chip-tuning in Appendix \ref{sec:finetune_baseline}.}

\paragraph{Chip-tuning excels previous baselines.}
It can be clearly observed that chip-tuning outperforms previously structured pruning baselines on almost every benchmark by a large margin, proving the capacity of our proposed model.
Meanwhile, while previously structured pruning baselines prune less than 30\% of the model parameters, chip-tuning is able to prune models by a higher ratio: 34.4\% for 7B models and 35.0\% for 13B models.

\paragraph{Linear chips are sufficient for classification.}
We also notice that the performance of linear chips is close to the performance of MLP chips, indicating that the features related to the input question may be mostly encoded linearly, and linear probing classifiers are enough for reading out these features.
Details of the difference will be demonstrated in Section \ref{ssec:layer_influence}. 

\paragraph{Optimal chips exhibit more potential.}
Finally, we gather the highest accuracy of all chips on each benchmark, notated as CT (max) in the table.
The pruning ratio and corresponding layer of optimal chips varies across different benchmarks and models (see Appendix \ref{sec:app_main_details} for details).
By choosing the optimal chip, chip-tuning could achieve even higher pruning ratios and performance.

\subsection{Pruning Multimodal Models}
We further evaluate whether chip-tuning could be applied to multimodal large language models (MLLMs) by pruning LLaVA-1.5 on image classification benchmarks.
Following the settings in \cite{zhang2024visually}, we train the chips for 500 epochs with a learning rate of $1\times 10^{-3}$, and set the batch size to 512. 

Table \ref{tab:mllm_comparison_all} demonstrates the results of pruning (see Appendix \ref{sec:app_mllm_details} for details).
Surprisingly, the original LLaVA models perform poorly on fine-grained image classification tasks, achieving an accuracy of near 0\% on Flowers102 and StanfordCars.
Providing the label set in the prompt could improve the accuracy, but the performance is still not satisfactory. 

In contrast, by tuning chips on the hidden states, we can achieve a decent accuracy while pruning the language model part of LLaVA.
This phenomenon indicates that the information essential for image classification is already contained in the hidden states of multimodal models, but the models have difficulty in correctly decoding them.
By adopting chip-tuning, we can extract related information before the final layer, and decode the information correctly.

We also notice that MLP chips perform extremely badly on StanfordCars, which may be caused by the large label set size of the dataset.

\begin{table*}[t]
	\centering
    \scalebox{0.9}{
        \begin{tabular}{ccccccccc}
		\toprule
        Model & Method & $\Delta$Params & BoolQ & Race-H & Race-M & C3 & MMLU & Avg. Score \\  
		\midrule
		\multirow{4}{*}{Llama2-7B} & Raw & - & 71.62 & 35.71 & 34.19 & 43.56 & 45.39 & 46.09 \\
		& LoRA & 8M & 87.37 & \textbf{81.59} & \textbf{86.56} & 83.83 & \textbf{54.80} & 78.83 \\
        & CT (Raw) & 0.5M & 79.05 & 47.91 & 53.69 & 48.93 & 44.89 & 54.89 \\
        & CT (LoRA) & 0.5M & \textbf{89.20} & 81.45 & 86.42 & \textbf{84.28} & 54.57 & \textbf{79.18} \\
		\midrule
		\multirow{4}{*}{Llama2-13B} & Raw & - & 82.39 & 57.95 & 60.38 & 47.51 & 55.00 & 60.65 \\
		& LoRA & 12.5M & 89.42 & 85.05 & 88.23 & \textbf{88.10} & \textbf{57.68} & \textbf{81.70} \\
        & CT (Raw) & 0.625M & 78.23 & 62.04 & 67.06 & 68.21 & 52.79 & 65.67 \\
        & CT (LoRA) & 0.625M & \textbf{90.09} & \textbf{85.22} & \textbf{88.58} & 87.81 & 55.51 & 81.44 \\
		\bottomrule
	\end{tabular}
    }
	\caption{Comparison between chip-tuning and finetuning with LoRA on the same training dataset. We attach a linear chip to the 20th layer of the 7B model and the 25th layer of the 13B model for classification. CT (Raw) and CT (LoRA) refer to adding linear chips to the raw model and the finetuned model, respectively.}
	\label{tab:lora_comparison}
\end{table*}

\subsection{Combination with Finetuning}
A critical difference between chip-tuning and the previous structured pruning method is that chip-tuning requires additional training data.
With these training data, we can also finetune the backbone language model to achieve better performance.
To better study the effectiveness of chip-tuning, we finetune models with the same data using LoRA~\cite{hu2021lora} and observe the performance gap between chip-tuning and finetuning.
We set rank $r = 16$ and LoRA alpha $\alpha = 32$\footnote{See Appendix \ref{sec:app_lora} for detailed settings.}.

Table \ref{tab:lora_comparison} shows the comparison results.
Finetuning the backbone model with LoRA could improve the performance on various benchmarks, and outperforms chip-tuning on the raw model as expected.
Nevertheless, we can perform chip-tuning on finetuned models, which will only lead to marginal performance loss and will even improve the performance on certain datasets.
These results clearly indicate that chip-tuning is compatible with traditional finetuning methods.

Considering that the target of probing is to read out relevant features from the internal representations of models, finetuning the model would help the backbone model develop better representations for the given classification task.
Thus, chips could benefit from the optimized input features and achieve better performance. 


\begin{figure*}
    \centering
    \subfloat[Llama-2-7B on MMLU.]{
    \includegraphics[width=0.48\textwidth]{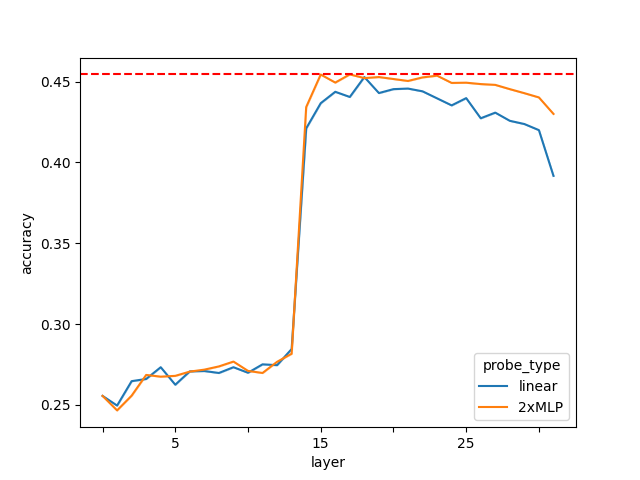}
    }
    \subfloat[Llama-2-13B on MMLU.]{
    \includegraphics[width=0.48\textwidth]{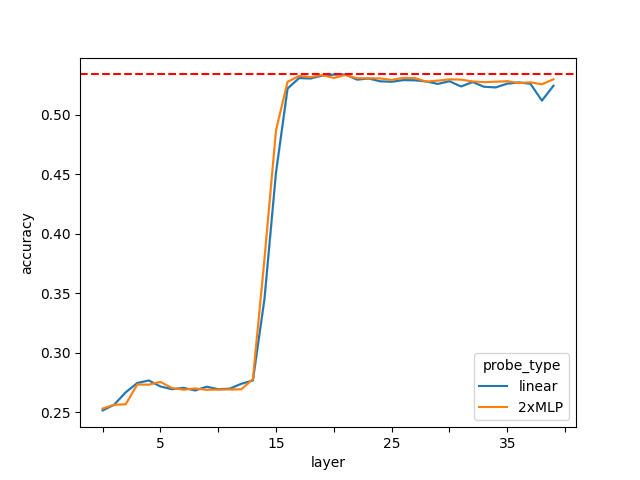}
    }
    \caption{The impact of pruning Llama2 models on MMLU by selecting chips on different layers.}
    \label{fig:layers_pruned}
\end{figure*}

\section{Analysis}
\subsection{Number of Pruned Layers}
\label{ssec:layer_influence}
Choosing different chips would change the number of pruned layers, and thus affect the classification performance.
We conduct experiments on the MMLU dataset with Llama-2 models, and Figure \ref{fig:layers_pruned} demonstrates the correlation between number of pruned layers and classification accuracy.

It can be clearly observed that the classification accuracy exhibits a drastic change on both datasets, increasing from random guess to a decent level, and then fluctuating within a relatively small range.
The change happens at around layer 18 for Llama-2-7B and happens at layer 20 for Llama-2-13B, which are at the position of about 50\% in the entire model.
It is also worth noticing that the best performance is not necessarily achieved on the last layer, especially for the Llama-2-7B model, which may be a hint that features in middle-layer representations serve better for classification.

\citet{stolfo2023mechanistic} proposes the theory that early layers in language models focus on \textit{gathering} and \textit{transmitting} information in the input text, while mid-late layers are involved in \textit{processing} the information and output the final answer.
The theory matches our findings: information relevant to the final answer is transmitted to the last token on intermediate layers, and the information is sufficient for solving the question.

We also find that the performance gap between linear chips and 2-layer MLP chips is not extremely significant.
On most layers, the two chips behave identically, especially for the 13B model.
The observable difference is that the performance of MLP chips is slightly more stable, changing in a smaller range on late layers.

\begin{figure}
    \centering
    \includegraphics[width=\linewidth]{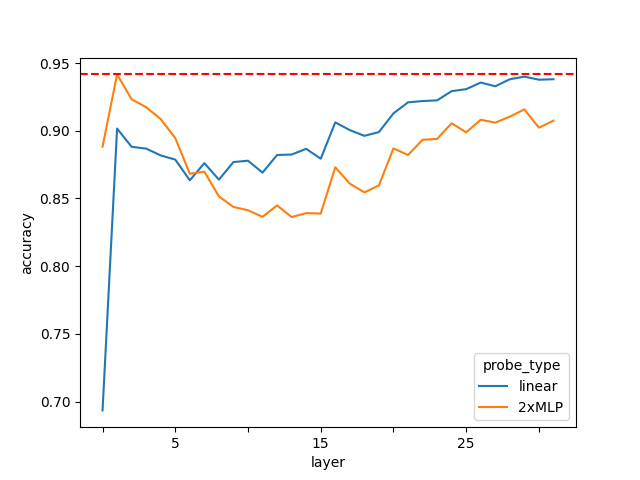}
    \caption{The impact of pruning LLaVA1.5-7B on Flowers102 by selecting chips on different layers. Different from the trends on NLP benchmarks, the trend does not exhibit a drastic change on certain layers.}
    \label{fig:mllm_prune_ratio}
\end{figure}

\begin{table*}[htbp]
	\centering
        \begin{tabular}{cccccccc}
		\toprule
        Model & Strategy & BoolQ & Race-H & Race-M & C3 & MMLU & Avg. Score \\  
		\midrule
		 \multirow{4}{*}{Llama2-7B} & Dense & 71.62 & 35.71 & 34.19 & 43.56 & 45.39 & 46.09 \\
        & Fixed & 79.05 & 47.91 & 53.69 & 48.93 & 44.89 & 54.89 \\
        & Validate & \textbf{79.48} & 49.43 & 54.53 & 53.93 & 44.53 & 56.38 \\
        & Optimal & \textbf{79.48} & \textbf{50.34} & \textbf{54.74} & \textbf{54.35} & \textbf{45.49} & \textbf{56.88} \\
		\bottomrule
	\end{tabular}
	\caption{Comparison between different chip selection strategies on Llama-2-7B.}
	\label{tab:chip_strategy}
\end{table*}

Multimodal models exhibit a different pattern.
As illustrated in Figure \ref{fig:mllm_prune_ratio}, chips on multimodal models could achieve high accuracy from early layers, and chips on late layers generally perform better than those on intermediate layers. 
The critical information for image classification is already contained in the image tokens from the first layer, which could lead to the difference.


\subsection{Chip Selection Strategy}
\label{ssec:chip_selection}
Aside from choosing chips on a fixed layer, there exist other strategies to achieve better performance.
We adopt three distinct strategies and evaluate them on Llama-2-7B:

\paragraph{Fixed} selects a fixed layer $l$ for all tasks ($l = 20$ for 7B models and $l = 25$ for 13B models), which is the strategy we use in main experiments.
\paragraph{Validate} constructs a small validation set consisting of 200 examples, and chooses the chip which performs best on the validation set.
\paragraph{Optimal} evaluates the performance of all chips, and selects the chip with the highest accuracy. This strategy reflects the upper bound of chip-tuning.

The experimental results are shown in Table \ref{tab:chip_strategy}.
Choosing chips according to the validation set generally achieves better performance than pruning the model on a fixed layer, but the pruning ratio may vary across different datasets.
While the \textbf{Optimal} strategy outperforms other strategies, the performance gap is not large,
The \textbf{Validate} strategy could achieve comparable results with \textbf{Optimal} accuracy, proving the robustness of chip-tuning.

\begin{figure}
    \centering
    \includegraphics[width=\linewidth]{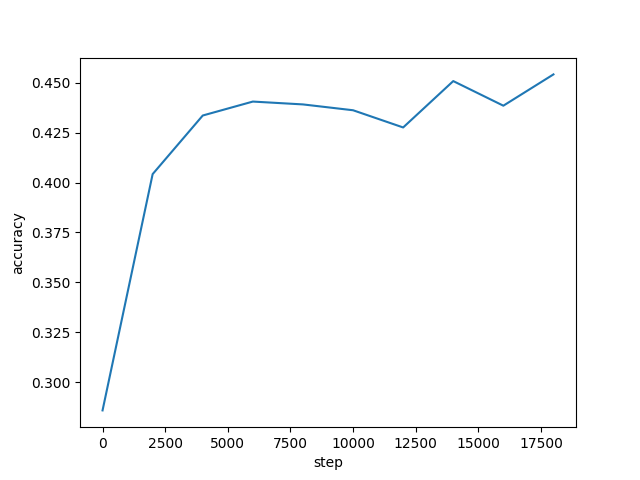}
    \caption{Analysis on the training dataset scale. We evaluate the performance of chip-tuning Llama-2-7B on MMLU every $2,000$ training steps. The overall accuracy rapidly increases until $6,000$ training steps, and continues to increase slightly afterward.}
    \label{fig:step_accuracy}
\end{figure}

\subsection{Impact of Training Dataset Scale}
Training data is a crucial component in model training.
Considering the scenario where training data is scarce, we test the performance of chip-tuning under different scales of the training dataset.

Figure \ref{fig:step_accuracy} shows the classification accuracy under different training dataset scales.
The accuracy rapidly increases before 6,000 training examples and reaches a plateau afterward.
Although the accuracy may drop at a certain time step, the figure still displays a pattern of slow increase after 6,000 training examples.
We draw the conclusion that a sufficient number of training data is essential for chips to converge, but further data could still bring subtle improvements.


\section{Conclusion}

In this paper, we propose chip-tuning, a structured pruning framework specialized for classification tasks.
Chip-tuning adopts probing classifiers to extract relevant features from intermediate layers of language models, and safely removes subsequent layers without affecting the selected classifier.
Experimental results on a variety of models and datasets demonstrate that chip-tuning surpasses previous baseline models on both performance and pruning ratio.
Chip-tuning performs well by selecting chips on a fixed layer, and could further achieve a pruning ratio of up to 50\% by selecting the optimal chip.

Meanwhile, we find that chip-tuning is also compatible with multimodal models and finetuned models.
Considering the simplicity of layer removal, chip-tuning shows its potential in deploying LLMs under practical scenarios.
We hope our work could inspire further research on efficient model pruning.

\section*{Limitations}
Based on the technique of probing, chip-tuning requires the backbone models to contain relevant features in their internal representations.
On tasks that the backbone models perform poorly, chip-tuning would not yield satisfactory results either.

Meanwhile, chip-tuning is designed mainly for classification tasks, which is the reason why we don't evaluate chip-tuning on datasets like HellaSwag that use perplexity-based evaluation methods.
Directly applying chip-tuning to generation tasks may lead to unexpected results, and generation-oriented chips remain to be explored in the future.


\bibliography{custom}

\clearpage

\appendix

\section{Datasets}
The properties of datasets we used are shown in Table \ref{tab:datasets}. We use these datasets according to their license and intended use.

\section{Details for Main Experiments}
\label{sec:app_main_details}
\begin{figure*}[htbp]
    \centering
    \subfloat[Llama-2-7B on BoolQ.]{
    \includegraphics[width=0.23\textwidth]{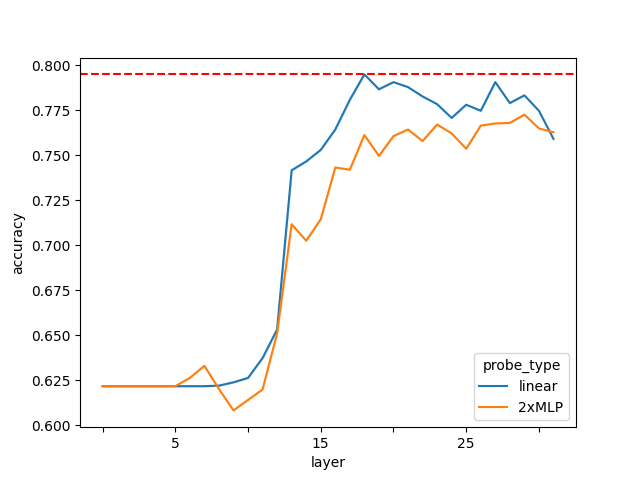}
    }
    \subfloat[Llama-2-13B on BoolQ.]{
    \includegraphics[width=0.23\textwidth]{figures/layer/accuracy_MMLU_l13.png}
    }
    \subfloat[Llama-2-7B on C3.]{
    \includegraphics[width=0.23\textwidth]{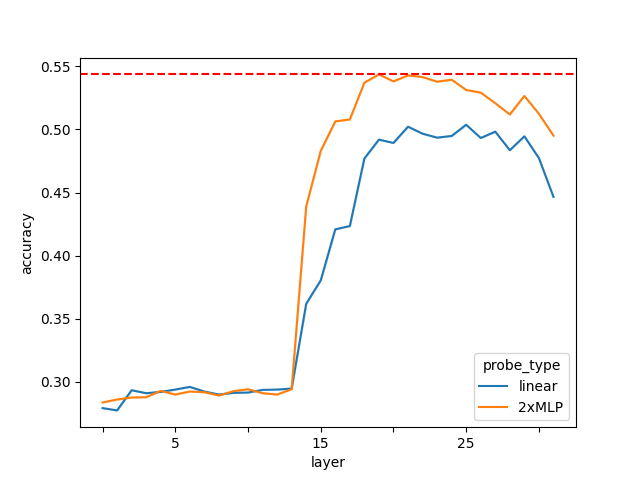}
    }
    \subfloat[Llama-2-13B on C3.]{
    \includegraphics[width=0.23\textwidth]{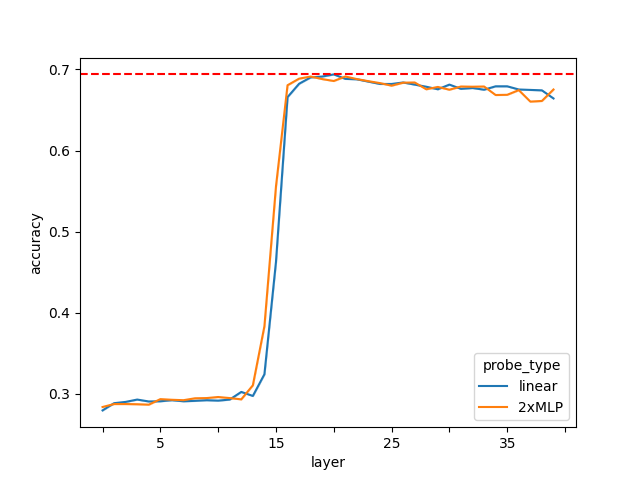}
    }
    \\
    \subfloat[Llama-2-7B on Race-H.]{
    \includegraphics[width=0.23\textwidth]{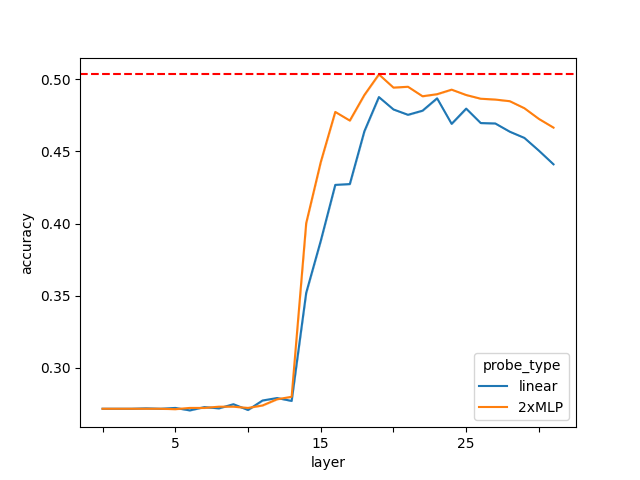}
    }
    \subfloat[Llama-2-13B on Race-H.]{
    \includegraphics[width=0.23\textwidth]{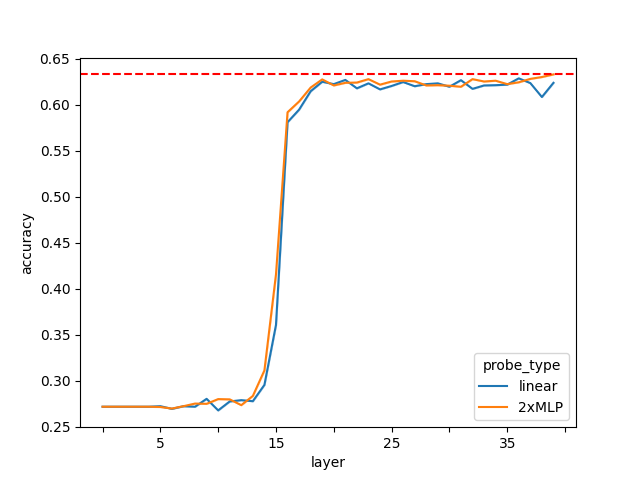}
    }
    \subfloat[Llama-2-7B on Race-M.]{
    \includegraphics[width=0.23\textwidth]{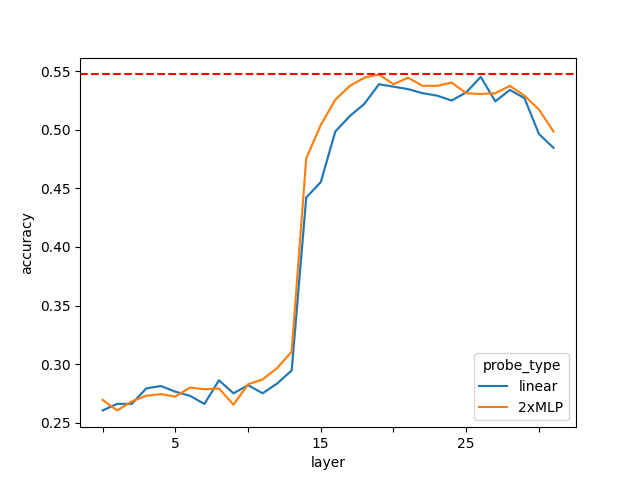}
    }
    \subfloat[Llama-2-13B on Race-M.]{
    \includegraphics[width=0.23\textwidth]{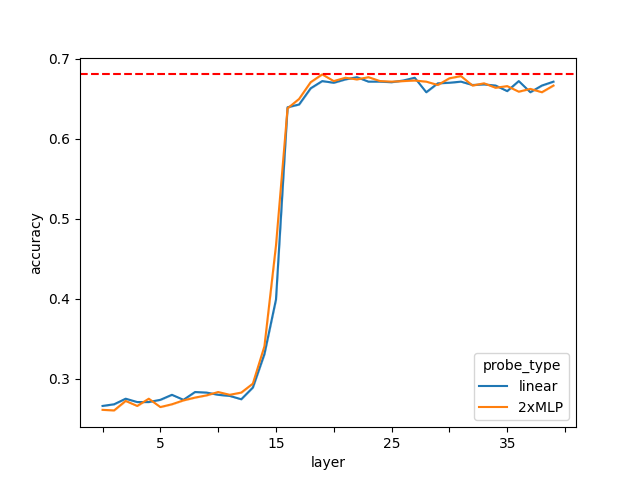}
    }
    \caption{The impact of pruning Llama2 models on BoolQ, C3, Race-H, and Race-M by selecting chips on different layers.}
    \label{fig:layers_pruned_details}
\end{figure*}

Figure \ref{fig:layers_pruned_details} shows how the performance changes with different number of layers pruned.
We can see that the optimal chip varies as the dataset changes.
However, pruning around layer 18 of the 7B model (about 40\%) and layer 20 of the 13B model (about 50\%) is generally acceptable.

We also notice that probing late layers of Llama-2-7B leads to worse results, which leaves the question of whether the 7B model "forgets" certain information on late layers.
The question remains to be explored in the future.

We record the layer on which chips show the best performance or highest pruning ratio in Table \ref{tab:best_layer}.
Notice that we define layer with the highest pruning ratio as the first layer after the drastic change in accuracy, which could be subjective.

We implement our code with the huggingface Transformers and Peft Python library.
Conducting chip-tuning on a 7B model or a 16-bit 13B model with 20,000 examples would take about 2 hours on a single NVIDIA A100 40GB GPU.

\section{Details for Multimodal Experiments}
\label{sec:app_mllm_details}
\begin{figure*}[htbp]
    \centering
    \subfloat[LLaVA1.5-7B on Flowers102.]{
    \includegraphics[width=0.31\textwidth]{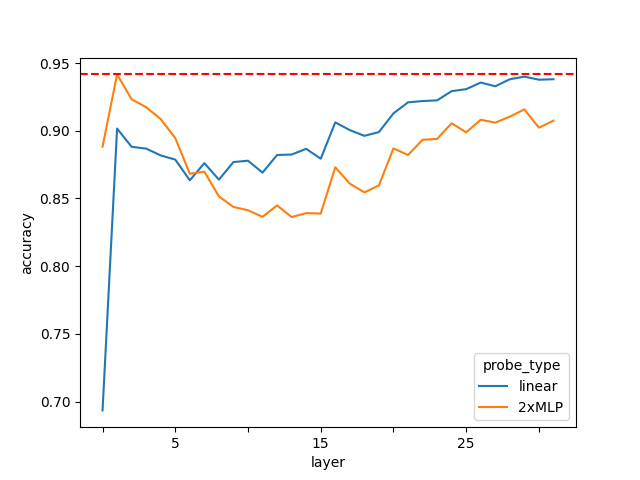}
    }
    \subfloat[LLaVA1.5-7B on StanfordCars.]{
    \includegraphics[width=0.31\textwidth]{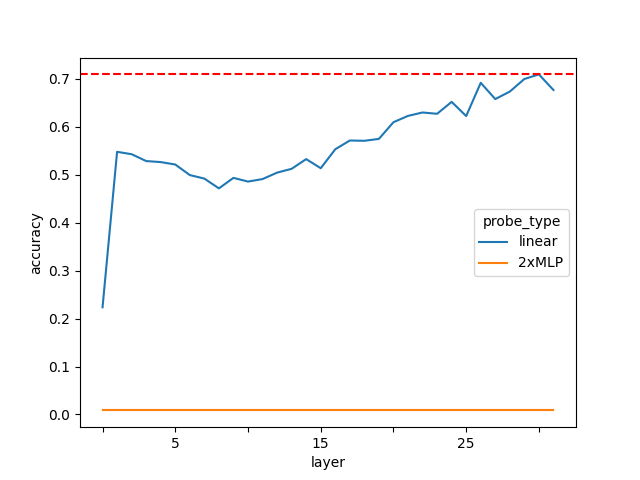}
    }
    \subfloat[LLaVA1.5-7B on Caltech101.]{
    \includegraphics[width=0.31\textwidth]{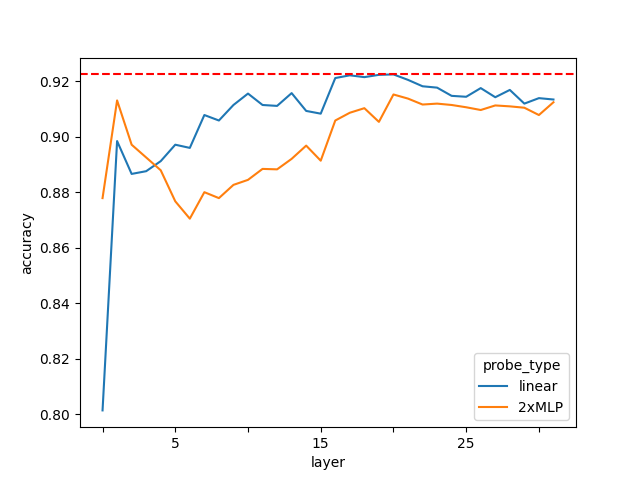}
    }
    \caption{The impact of pruning LLaVA1.5-7B on Flowers102, StanfordCars, and Caltech101 by selecting chips on different layers.}
    \label{fig:mllm_layers_pruned_details}
\end{figure*}

Figure \ref{fig:mllm_layers_pruned_details} shows how the performance changes by pruning LLaVA1.5-7B.
Different from text datasets, the optimal chip for image classification typically appears on late layers, while chips on early layers also exhibit decent accuracy.
Surprisingly, 2-layer MLP chips fail to predict the class of images on StanfordCars.
This may be a result of the larger class label set size (196) compared with Flowers102 (102) and Caltech101 (101).

\section{Details for LoRA Experiments}
\label{sec:app_lora}
Table \ref{tab:app_lora_params} shows the experimental settings for LoRA experiments.

\section{Experiments on Llama3}
\label{sec:app_llama3}

We evaluate chip-tuning on Llama3-8B-Instruct~\cite{llama3modelcard}, one of the up-to-date LLMs.
We prune the model to layer 22 in experiments.

The experimental results in Table \ref{tab:llama3} are similar to those in Table \ref{tab:lora_comparison}: applying chip-tuning on Llama3 has minimal impact on classification accuracy, proving that chip-tuning is compatible with Llama3.
The optimal performance of chips even outperforms the finetuned LoRA models.

\section{Finetuning Pruned Baseline Models}
\label{sec:finetune_baseline}
For fair comparisions, we finetune the pruned baseline models on the training set of each benchmark to see how they perform with the same data provided.
We use Llama2-7B as the backbone model, and finetune LLMPruner~\cite{ma2023llm} and SliceGPT~\cite{ashkboos2024slicegpt} under their default LoRA settings.
We do not train LaCo and ShortGPT as we cannot find their official code.

Both baseline models are trained with at most 20,000 training data same to these chip-tuning used on each benchmark.
The accuracy of finetuned baseline models are obtained by selecting the choice token (for example, "A", "B", "C", "D" for 4-choice problems) with the highest generation probability, as free-form generation would yield unexpected results.

Table \ref{tab:finetune_baseline} shows the result of finetuning LLMPruner and SliceGPT with the same data used by chip-tuning.
While the finetuned versions achieve higher accuracy than the original version, we can clearly see that chip-tuning greatly outperforms both baselines, further proving the effectiveness of chip-tuning.

\begin{table}[htbp]
    \centering
    \begin{tabular}{c c}
    \hline
        Parameter & Value \\
    \hline
        learning rate & $1\times 10^{-5}$ \\
        weight decay & 0.01 \\
        r & 16 \\
        $\alpha$ & 32 \\
        batch size & 1 \\
        epoch & 1 \\
    \hline
    \end{tabular}
    \caption{Parameters for LoRA training.}
    \label{tab:app_lora_params}
\end{table}

\begin{table}[htbp]
	\centering
    \scalebox{0.7}{
        \begin{tabular}{cccccc}
		\toprule
        Model & BoolQ & Race-H & Race-M & C3 & MMLU \\  
		\midrule
		Llama2-7B & 18/17 & 19/19 & 19/17 & 19/18 & 17/15  \\
		Llama2-13B & 38/18 & 39/18 & 19/16 & 20/17 & 21/16 \\
		Baichuan2-7B & 21/18 & 30/19 & 30/19 & 20/19 & 24/19 \\
		Baichuan2-13B & 38/18 & 36/22 & 35/22 & 27/22 & 22/21 \\
		\bottomrule
	\end{tabular}
 }
	\caption{The corresponding layer of chips with the best performance or highest pruning ratio on each dataset. The format of each cell in the table is (layer with best performance / layer with highest pruning ratio).}
	\label{tab:best_layer}
\end{table}

\begin{table*}
    \centering
    \scalebox{0.8}{
    \begin{tabular}{c c c c}
    \hline
        Dataset & Link & Train Split & Eval Split \\
    \hline
        BoolQ & https://huggingface.co/datasets/google/boolq & train & validation \\
        Race & https://huggingface.co/datasets/ehovy/race & train & test \\
        C3 & https://huggingface.co/datasets/dataset-org/c3 & train & validation \\
        MMLU & https://huggingface.co/datasets/cais/mmlu & auxiliary\_train & test \\
        Flowers102 & https://huggingface.co/datasets/dpdl-benchmark/oxford\_flowers102 & train+validation & test \\
        StanfordCars & https://huggingface.co/datasets/tanganke/stanford\_cars & train & test \\
        Caltech101 & https://huggingface.co/datasets/dpdl-benchmark/caltech101 & train & test \\
    \hline
    \end{tabular}
    }
    \caption{Dataset details.}
    \label{tab:datasets}
\end{table*}

\begin{table*}[htbp]
	\centering
        \begin{tabular}{ccccccccc}
		\hline
        Model & Method & $\Delta$Params & BoolQ & Race-H & Race-M & C3 & MMLU & Avg. Score \\  
		\hline
		\multirow{4}{*}{Llama3-8B} & Raw & - & 57.77 & 80.87 & 85.24 & 86.82 & 64.01 & 74.94 \\
		& LoRA & 8M & 87.16 & 80.25 & 90.46 & 84.93 & 66.20 & 81.8 \\
        & CT (Raw) & 0.5M & 76.73 & 81.39 & 86.00 & 88.86 & 64.26 & 79.45 \\
        & CT (LoRA) & 0.5M & 87.03 & 81.36 & 90.95 & 83.75 & 66.31 & 81.88 \\
        & CT (Max) & 0.5M & \textbf{90.83} & \textbf{88.02} & \textbf{91.07} & \textbf{93.71} & \textbf{66.37} & \textbf{86.00} \\
		\hline
	\end{tabular}
	\caption{Comparison between chip-tuning and finetuning with LoRA on Llama3-8B. CT (Raw) and CT (LoRA) refer to adding linear chips to the raw model and the finetuned model on layer 22, respectively. CT (Max) refers to the best performance of chips on the finetuned model.}
	\label{tab:llama3}
\end{table*}

\begin{table*}[htbp]
	\centering
        \begin{tabular}{cccccccc}
		\hline
        Model & Method & BoolQ & Race-H & Race-M & C3 & MMLU & Avg. Score \\  
		\hline
		\multirow{4}{*}{Llama2-7B} & Raw & 71.62 & 35.71 & 34.19 & 43.56 & 45.39 & 46.09 \\
        & LLMPrun. & 55.20 & 22.56 & 22.35 & 25.64 & 23.33 & 29.82 \\
		& SliceGPT & 38.32 & 21.07 & 21.66 & 39.78 & 28.92 & 29.95 \\
        & CT (Raw) & 79.05 & 47.91 & 53.69 & 48.93 & 44.89 & 54.89 \\
        \cline{2-8}
        & LoRA & 87.37 & \textbf{81.59} & \textbf{86.56} & 83.83 & \textbf{54.80} & 78.83 \\
        & LLMPrun.+LoRA & 60.49 & 23.67 & 25.56 & 17.84 & 26.60 & 30.83 \\
        & SliceGPT+LoRA & 65.72 & 71.76 & 76.81 & 60.27 & 43.06 & 63.52 \\
        & CT (LoRA) & \textbf{89.20} & 81.45 & 86.42 & \textbf{84.28} & 54.57 & \textbf{79.18} \\
		\hline
	\end{tabular}
	\caption{Comparison between chip-tuning and finetuning pruned baseline models. We only report the result of finetuning LLMPruner and SliceGPT, as LaCo and ShortGPT do not provide their official code.}
	\label{tab:finetune_baseline}
\end{table*}

\end{document}